\documentclass[runningheads]{llncs}

\usepackage[T1]{fontenc}
\usepackage{graphicx}

\usepackage[utf8]{inputenc}
\usepackage{hyperref}
\usepackage{comment}
\usepackage{todonotes}
\usepackage{wrapfig}
\usepackage{amsfonts}
\usepackage[misc]{ifsym}
\usepackage{multicol}
\usepackage{rotating}
\usepackage{array}
\usepackage{lipsum}

\title{A Survey on Knowledge Graph-based Methods for Automated Driving}

\author{
Juergen Luettin\inst{1} \and
Sebastian Monka\inst{1} \and
Cory Henson \inst{2} \and
Lavdim Halilaj\inst{1} 
}

\institute{Robert Bosch GmbH - Corporate Research, Renningen, Germany \\
\and
Robert Bosch LLC - Bosch Research and Technology Center, Pittsburgh, PA, USA \\
\email{\{firstname.lastname\}@bosch.com}
}

\begin{document}
\maketitle
\thispagestyle{empty}
\pagestyle{empty}
\let\thefootnote\relax\footnotetext{Preprint, under review.}

\begin{abstract}
Automated driving is one of the most active research areas in computer science. 
Deep learning methods have made remarkable breakthroughs in 
machine learning in general and in automated driving (AD) in particular.
However, there are still unsolved problems 
to guarantee reliability and safety of automated systems,
especially to effectively incorporate 
all available information and knowledge in the driving task.
Knowledge graphs (KG) have recently gained significant attention 
from both industry and academia for applications that benefit by exploiting 
structured, dynamic, and relational data.
The complexity of graph-structured data with complex relationships and 
inter-dependencies between objects has posed significant challenges
to existing machine learning algorithms.
However, recent progress in knowledge graph embeddings and graph neural networks 
allows to applying machine learning to graph-structured data. 
Therefore, we motivate and discuss the potential benefit of KGs applied to the main tasks of AD including 1) ontologies 2) perception,  3) scene understanding, 4) motion planning, and 5) validation.
Then, we survey, analyze and categorize ontologies and KG-based approaches for AD.
We discuss current research challenges and propose promising future research directions for KG-based solutions for AD.
\end{abstract}


\section{Introduction}
\vspace{-0.2cm}
The first successful AD vehicle was demonstrated in
the 1980s~\cite{Dickmanns2005DynamicMM}.
However, despite remarkable progress, fully AD has not been realized to date.
One unsolved problem is that AD vehicles must 
be able to drive safely in situations that have not been seen before in the training data.
Moreover, AD systems must consider the strong safety requirements specified in ISO 26262~\cite{ISO26262}, which states that the behavior of the components needs to be fully specified and that each refinement can be verified with respect to its specification.
verified.
Adhering to this standard is a prerequisite for automated driving systems.

Deep learning (DL)~\cite{Hinton2006AFL,Schmidhuber2015DeepLI,Goodfellow2015DeepL}
has made remarkable breakthroughs with significant
impact on the performance of AD systems.
However, DL methods do not provide information to 
adequately understand what the network 
has learned and thus are hard to interpret and validate~\cite{Burton2017MakingTC,Herrmann2022UsingOF}.
In safety-critical applications, this is a major drawback.
The concept of explainable AI (XAI)~\cite{Arrieta2020ExplainableAI}
was therefore introduced.

Moreover, the performance of DL methods is heavily dependent on the availability of suitable training data.
When the testing environment deviates from the distribution of the training data, DL methods tend to produce unpredictable and critical errors.
Whereas driving is governed by traffic laws and typical driver behaviors that
represents a crucial knowledge source, traditional DL methods cannot easily incorporate such explicit knowledge.
We argue that KGs are well suited to address all of these drawbacks.

\begin{figure*}[tb]
	\centering
	\includegraphics[width=1.0 \textwidth]{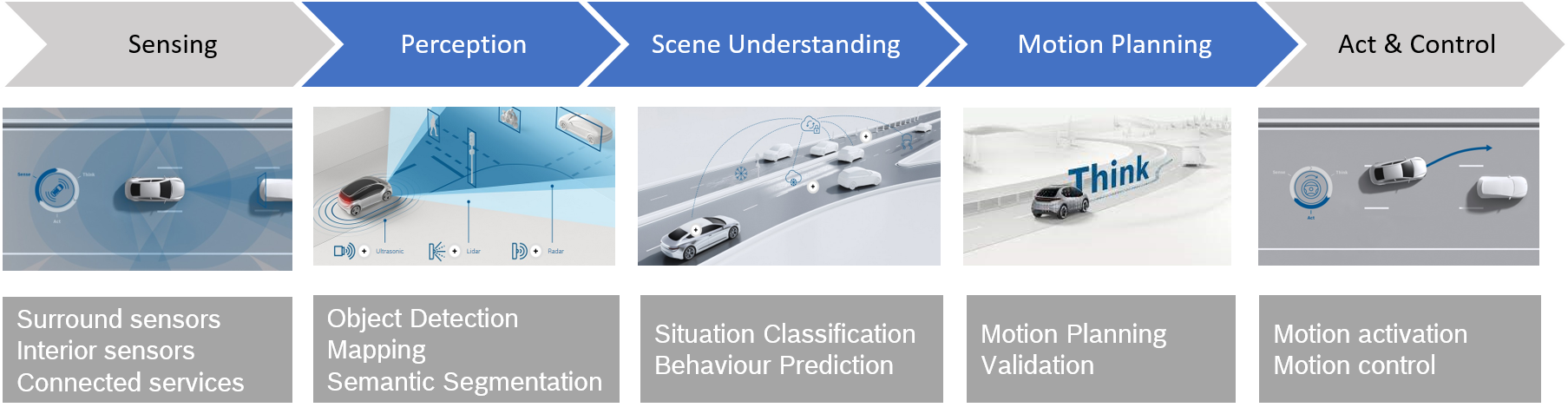}
	\caption{\textbf{Standard components of an AD system}, modified from~\cite{Kiran2022DeepRL}.  }
	\vspace{-0.3cm}
	\label{fig:AD_components}
\end{figure*}

The use of graphs to represent knowledge has been researched for a long time.
The term knowledge graph (KG) was popularized  with the announcement of the Google Knowledge Graph~\cite{Singhal2012Introducing}.
A graph-based representation has several advantages over alternative approaches to represent knowledge.
Graphs represent a concise and intuitive abstraction with edges representing the 
relations that exist between entities.
Special graph query languages allow to find complex arbitrary-length paths in the graph. 
Recently, methods to apply machine learning directly on graphs have generated new opportunities to use KGs in data-based applications~\cite{Wang2017KnowledgeGE}.
Figure~\ref{fig:AD_components} shows the standard components of an AD system together with their sub-tasks. 
In this survey, we address KG-based approaches for the components \textit{Perception, Scene Understanding, and Motion Planning}.

\vspace{-0.2cm}

\section{Preliminaries}
\vspace{-0.2cm}

We first describe the basic terminology relevant in the context of this survey as well as insights of related work regarding generic principles of joint usage of knowledge graphs and machine learning pipelines.

\subsection{Knowledge Graphs \& Ontologies}

Knowledge graphs are means for structuring facts, with entities connected via named relationships. 
Hogan et al.~\cite{Hogan2021Knowledge} define a KG as "a graph of data with the objective of accumulating and conveying real-world knowledge, where nodes represent entities and edges are relations between entities".
Entities can be real-world objects and abstract concepts, relationships represent the relation between entities, and semantic descriptions of entities and their relationships contain types and properties with a well-defined meaning.
Knowledge can be expressed in a factual triple in the form of (head, relation, tail) or (subject, predicate, object) 
under the Resource Description Framework (RDF), for example, (Albert Einstein, WinnerOf, Nobel Prize). 
A KG is a set of triples $G = {H, R, T}$, where $H$ represents a set of entities $E$, $T \subseteq E \times L $, a set of entities and literal values, and $R$ set of relationships connecting $H$ and $T$.


\subsection{Models}

A graph model is a model which structures the data, including its schema and/or instances in the forms of graphs, and the data manipulation is realized by graph-based operations and adequate integrity
constraints\cite{DBLP:books/sp/18/AnglesG18}.
Each graph model has its own formal definition based on the mathematical foundation, which can vary according to different characteristics, for instance, directed vs undirected, labeled vs unlabeled, etc.
The most basic model is composed of labeled nodes and edges, easy to comprehend but inappropriate to encapsulate multidimensional information.
Other graph models allow for representation of information utilizing complex relationships in the form of hypernodes or hyperedges.

Here we discuss three common graph models used in practice to represent data graphs.

\subsubsection{Directed Labeled Graphs}
A directed labeled graph is comprised of a set of nodes and a set of edges connecting those nodes, labeled based on a specific vocabulary. 
The direction of the edge of two paired nodes is important, which clearly distinguishes between the start node and the end node.
This enables the organization of information in an intuitive way via the utilization of binary relationships. 

\subsubsection{Hyper-relational Graphs}
A hyper-relational graph is also a labeled directed multigraph where each node and edge might have a number of associated key-value pairs \cite{DBLP:journals/access/AnglesTT20}. 
Internally, nodes and edges are annotated according to a chosen vocabulary and have unique identifiers, making them a flexible and powerful form of modeling for graph analysis with weighted edges. 

\subsubsection{Hypergraphs}
Hypergraphs enhance the definition of binary edges by allowing modeling of multiple and complex relationships. 
On the other hand, hypernodes modularize the notion of node, by allowing nesting graphs inside nodes.
In addition, the notion of a hyperedge enables the definition of n-ary relations between different concepts.

A knowledge graph can be based on any such graph model utilizing nodes and edges as a fundamental modeling form.
KGs are essentially composed of two main components: 1) schemas a.k.a. ontologies; and 2) the actual data modeled according to the given ontologies.
In philosophy, an ontology is considered as a systematic study of things, categories, and their relations within a particular domain. 
In computer science, on the other hand, ontologies are defined as a formal and explicit specification of a shared conceptualization~\cite{Studer1998Knowledge}.
They enable conceptualization of knowledge for a given domain and support common understanding between various stakeholders.
Thus, ontologies are a crucial component in tackling the semantic heterogeneity problem and enabling interoperability in scenarios where different agents and services are involved.

\subsection{Knowledge Representation Learning}

While most symbolic knowledge is encoded in graph representation, conventional machine learning methods operate in vector space.
Using a \emph{knowledge graph embedding method} (KGE-Method), a KG can be transformed into a \emph{knowledge graph embedding} (KGE), a representation where entities and relations of a KG are mapped into low-dimensional vectors.
The KGE captures semantic meanings and relations of the graph nodes and reflects them by locality in vector space~\cite{Wang2017KnowledgeGE}.
KGEs are originally used for graph-based tasks such as node classification or link prediction, but have recently been applied to tasks such as object classification, detection, or segmentation.
As defined in~\cite{DBLP:journals/corr/abs-2005-03675}, graph embedding algorithms can be clustered into unsupervised and supervised methods.

\textit{Unsupervised KGE-Methods} form a KGE based on the inherent graph structure and the attributes of the KG.
One of the earlier works~\cite{Nickel2016ARO} 
focused on statistical relational learning.
Recent surveys~\cite{Cai2018ACS,Wang2017KnowledgeGE,Goyal2018GraphET,Ji2021ASO} categorize unsupervised KGE-Methods based on their \emph{representation space} (vector, matrix, and tensor space), the \emph{scoring function} (distance-based, similarity-based), the \emph{encoding model} (linear/bilinear models, factorization models, neural networks), and the \emph{auxiliary information} (text descriptions, type constraints).

\textit{Supervised KGE-Methods} learn a KGE to best predict additional node or graph labels.
Forming a KGE by using task-specific labels for the attributes, the KGE can be optimized for a particular task while retaining the full expressivity of the graph.
Common supervised KGE-Methods are based on \emph{graph neural networks} (GNNs)~\cite{1555942}, an extension of DL networks that can directly work on a KG.
For scalability reasons and to overcome challenges that arise from graph irregularities, various adaptations have emerged such as \emph{graph convolutional networks} (GCN)~\cite{Kipf2017SemiSupervisedCW} and \emph{graph attention networks} (GAT)~\cite{DBLP:conf/iclr/VelickovicCCRLB18}.

Several surveys focusing on different research topics in AD have been published,
including 
\textit{computer vision}~\cite{Janai2020ComputerVF},
\textit{object detection}~\cite{Arnold2019ASO,Gouidis2020ARO},
\textit{DL based scene understanding}~\cite{Liu2019DeepLF,Guo2021ASO,Grigorescu2020ASO,Huang2020Survey},
\textit{DL based vehicle behavior prediction}~\cite{Mozaffari2019DeepLV},
\textit{deep reinforcement learning}~\cite{Kiran2022DeepRL},
\textit{lane detection}~\cite{Zhu2017OverviewOE,Narote2018ARO,Tang2021ARO},
\textit{semantic segmentation}~\cite{Lateef2019SurveyOS,Minaee2021ImageSU}, and     \textit{planning and decision making}~\cite{Gonzlez2016ARO,Paden2016ASO,Schwarting2018PlanningAD,Badue2021SelfDrivingCA,Claussmann2017ASO}.
More recently, ontologies and KGs have gained interest for knowledge-infused learning approaches.
Monka et al.~\cite{Monka2022ASO} provided a survey about visual transfer learning using KGs.
However, we did not find a survey that cover the use of KGs applied to AD.


\section{Ontologies for Automated Driving}
\vspace{-0.2cm}

\begin{figure*}[h]
\includegraphics[width=1\textwidth]{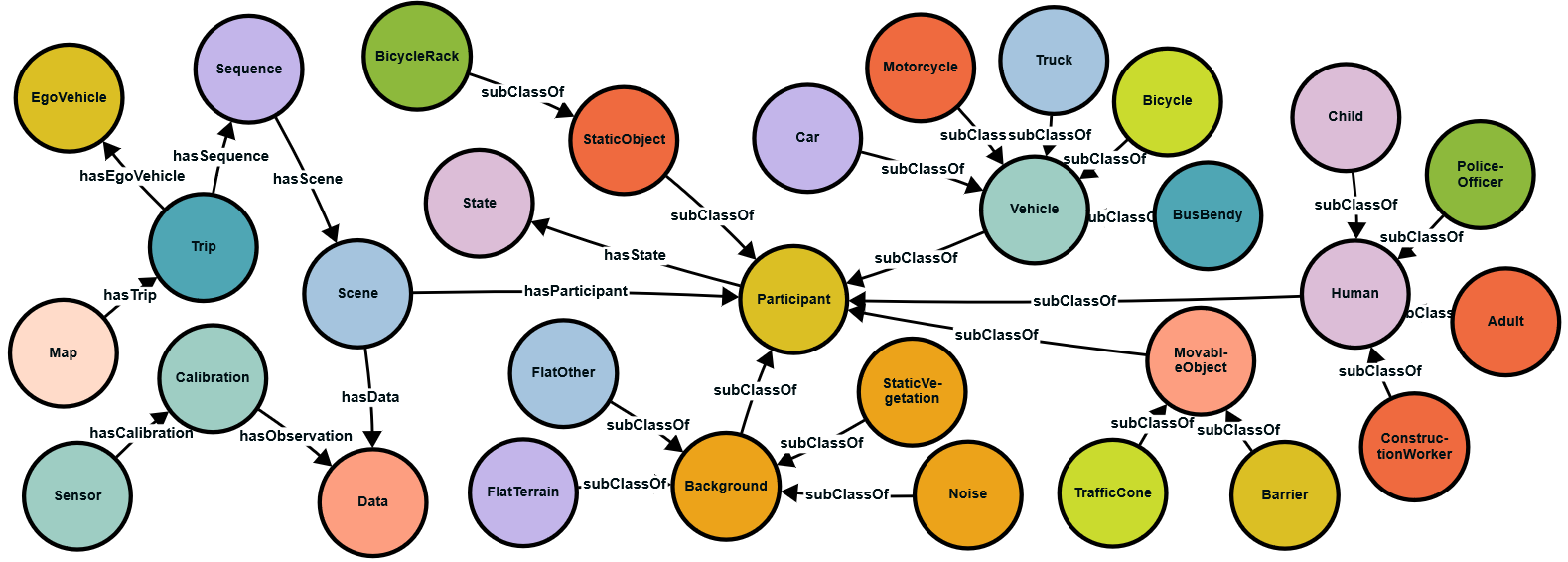}
\caption{\textbf{Scene Ontology}. Excerpt of an ontology to describe  information of a driving scene~\cite{Halilaj2022Knowledge}}
\vspace{-0.3cm}
\label{fig:nuSceneOnto}
\end{figure*}

Several ontologies have been developed to model relevant knowledge in the automotive domain.
They cover elements such as vehicle, driver, route, and scenery, including their spatial and temporal relationships.
The authors in~\cite{Geyer2014ConceptAD,DBLP:conf/itsc/UlbrichMRSM15} propose a consolidated definition and taxonomy for AD terminology.
The goal is to establish a standard and consistent terminology and ontology. 
Additionally, there exist well-known ontologies such as \emph{DBpedia}~\cite{DBLP:conf/semweb/AuerBKLCI07}, \emph{Schema.org}~\cite{DBLP:journals/cacm/GuhaBM16} and \emph{SOSA}~\cite{Janowicz2019SOSAAL} which contain concepts related to the automotive domain described from a more generic perspective.
Fig.~\ref{fig:nuSceneOnto} illustrates main concepts such as \emph{Scene}, \emph{Participant} and \emph{Trip} with sub-categories and relationships.
In the following, we categorize and describe the surveyed ontologies considering their primary focus.

\vspace{-0.2cm}

\subsubsection{Vehicle Model}

An ontology modeling the main concepts of vehicles, such as vehicle type, installed components and sensors, is described in~\cite{Zhao2015CoreOF}. 
The work in~\cite{DBLP:conf/semweb/KlotzTWB18} focuses on representing sensors, their attributes, and generated signals to increase data interoperability.

\subsubsection{Driver Model}
Combined approaches for managing driving related information by a model representing the driver, the vehicle, and the context are described in~\cite{Hina2017OntologicalAM,Feld2011TheAO}. 
An ontology for modeling driver profiles based on their demographic and behavioral aspects is proposed by~\cite{sarwar2019context}. 
It is used as an input for a hybrid learning approach for categorizing drivers according to their driving styles.

\subsubsection{Driver Assistance}

An  ontology for driver assistance  is described  
in~\cite{Kannan2010AnID}. 
It serves as a common basis for domain understanding, decision-making, and information sharing. 
Feld et al.~\cite{Feld2011TheAO} presented an ontology focusing on human-machine-interfaces and inter-vehicle communication.
Huelsen et al.~\cite{Hlsen2011TrafficIS,Hlsen2011AsynchronousRF} developed an ontology for traffic intersections to reason about the right-of-way including information about traffic signs and traffic lights.
Related approaches can be found in~\cite{Fuchs2008AMF,Gutirrez2014AgentbasedFF,DBLP:conf/hci/LilisZPAS17,Ryu2018ContextAwarenessBD}.

\subsubsection{Route Planning}
Schlenoff et al.~\cite{Schlenoff2003UsingOT} explored the use of ontologies to improve the capabilities and performance of on-board route-planning.
An ontology-based method for modeling and processing map data in cars was presented by Suryawanshi et al.~\cite{Suryawanshi2019AnOM}.
Kohlhaas et al.~\cite{Kohlhaas2014SemanticSS} introduced a modeling approach for the semantic state space for maneuver planning.

\begin{table*}[tb]
  \caption{\textbf{Automotive ontologies}. Relevant ontologies split into different categories considering their main focus and concepts.}
  \label{tab:AutomotiveOntologies}
  \small
	\centering
	\begin{tabular}%
	{>{\raggedright\arraybackslash}p{1.6cm}
	>{\raggedright\arraybackslash}p{3.5cm}
	>{\raggedright\arraybackslash}p{5.3cm}
	>{\raggedright\arraybackslash}p{1.4cm}}
	\hline
  \textbf{Focus}	  &  \textbf{Scope}  &
  \textbf{Main Concepts} &
  \textbf{Ref.}   \\
	\hline 
    Generic & Vehicle, Driver, Scene
    & Vehicle, Engine Spec., Usage Type, Driving Wheel, Configuration, Manufacturer, Sensor, Scene, Time & \cite{DBLP:conf/semweb/AuerBKLCI07,DBLP:journals/cacm/GuhaBM16,Janowicz2019SOSAAL}  \\
 
    Vehicle & Vehicle, Categories, Parts
 & Vehicle, Sensors, Actuators, Signals, Taxonomy, Speed, Acceleration & \cite{DBLP:conf/semweb/KlotzTWB18,Zhao2015CoreOF} \\
 
    Driver Model & Driver, Characteristics, Objectives 
  & Ability, Physiological \& Emotional State, Driving Style, Preferences, Behavior & \cite{Feld2011TheAO,Hina2017OntologicalAM,sarwar2019context} \\
  
   Driver Assistance & Driver Machine Interactions, Information Exchange 
  & Assistance Functions, Interaction Module, Exchange Interfaces  & \cite{Feld2011TheAO,Fuchs2008AMF,Gutirrez2014AgentbasedFF}
  \cite{Hlsen2011TrafficIS,Hlsen2011AsynchronousRF,Kannan2010AnID} 
  \cite{DBLP:conf/hci/LilisZPAS17,Ryu2018ContextAwarenessBD} \\
 
    Routing & Road Geometry,  Road Network, Intersections
 & Road Part, Road Type, Lane, Junction, Traffic Sign, Obstacles  &  \cite{Kohlhaas2014SemanticSS,Schlenoff2003UsingOT,Suryawanshi2019AnOM} \\

    Context Model & Scenario, Event, Situation
 & Road Participants, Static and Dynamic Environment, Driving Maneuvers, Temporal Relations & \cite{Buechel2017OntologybasedTS,Feld2011TheAO,Gelder2020OntologyFS}
 \cite{Halilaj2021AKnowledge,DBLP:conf/semweb/HensonSTK19,DBLP:conf/kes/MohammadKHS15} 
 \cite{DBLP:conf/fusion/PollardMN13,Ulbrich2014GraphbasedCR}  \\
 
    Cross-Cutting & Automation Level, Driving Task, Risk, Abstraction Layer
 & Automation Mode, Longitudinal \& Lateral Control, Risk Estimation, Autonomy and Communication & 
 \cite{LEROY2020212,DBLP:conf/fusion/PollardMN13,DBLP:conf/ice-itmc/SchaferKCG17}
 \cite{Scholtes20216LayerMF,Westhofen2022UsingOF,ASAMOpenScenario}
 \cite{ASAMOpenX,Urbieta2021DesignAI,Halilaj2022Knowledge}  \\
  
	\hline
	\end{tabular}
	\vspace{-0.3cm}
\end{table*}

\subsubsection{Context Model}

Ulbrich et al.~\cite{Ulbrich2014GraphbasedCR} described an ontology for context representation and environmental modelling for AD. 
It contains different layers including a metric layer of lane context, traffic rules, object, and situation description.
Buechel et al.~\cite{Buechel2017OntologybasedTS} proposed a modular framework for traffic regulation based decision-making in AD.
The traffic scene is represented by an ontology and includes knowledge about traffic regulations.
Henson et al.~\cite{DBLP:conf/semweb/HensonSTK19} presented an ontology-based method for searching scenes in AD datasets. 
Similar approaches representing the context in a driving scenario are shown in~\cite{Feld2011TheAO,Gelder2020OntologyFS,Halilaj2021AKnowledge,DBLP:conf/kes/MohammadKHS15}.
Ontologies have also been used for context-dependent recommendation tasks~\cite{Werner2021RETRA,DBLP:conf/sac/HalilajLRAD21}.

\subsubsection{Cross-Cutting}
An ontology describing the levels of automated driving systems, ranging from fully manual to fully automated was presented in~\cite{DBLP:conf/fusion/PollardMN13}.
In line with this,~\cite{DBLP:conf/ice-itmc/SchaferKCG17} analyses crucial questions of driving tasks and maps them to the level of driving automation. 
The interaction risks between human driven and automated vehicles investigated in~\cite{LEROY2020212} are based on five main components: obstacle, road, ego-vehicle, environment, and driver.
An informal but comprehensive 6-layer model for a structured description and categorization of urban scenes was describe in~\cite{Scholtes20216LayerMF}. 
This was further developed into the Automotive Urban Traffic Ontology ({A.U.T.O.})~\cite{Westhofen2022UsingOF}. The Automotive Global Ontology({AGO})~\cite{Urbieta2021DesignAI} focuses on semantic labeling and testing use cases. 
Ontologies representing data structures of AD datasets have been presented in~\cite{Halilaj2022Knowledge}.
Several standards aim to develop ontologies to provide a foundation 
of common definitions, properties, and relations for central concepts, 
e.g. {ASAM} Open{S}cenario~\cite{ASAMOpenScenario} and {ASAM} Open{X}~\cite{ASAMOpenX}.

A summary of automotive ontologies with scope and main concepts is shown in Table.~\ref{tab:AutomotiveOntologies}.
It can be seen that many ontologies cover only specific concepts and use cases but other are more complete and provide a more comprehensive coverage for all AD components and tasks.
Since one of the main benefits of ontologies is re-usability, these ontologies can be re-used and extended for various scenarios.



\section{Knowledge Graphs applied to Automated Driving}
\vspace{-0.2cm}
In this section, we provide an overview of KG-based approaches and categorize them based on their respective components and tasks in the AD pipeline as shown in Figure~\ref{fig:AD_components}. 
We consider approaches that use ontologies or KGs as well as approaches that combine KGs with machine learning for AD tasks.

\subsection{Object Detection} 
Object detection in AD includes the detection of traffic participants such as vehicles, pedestrians, road markings, traffic signs, and others. Typical sensors used are video cameras, RADAR and LiDAR.

Scene graphs are a relatively recent technique to semantically describe and represent objects and relations in a scene~\cite{Johnson2015ImageRU}. Much research is targeting the generation of scene graphs~\cite{Chang2021SceneGA} which can be divided into two types.
The first, bottom up approach, consists of object detection followed by pairwise relationship recognition. The second, top-down approach, consists of joint detection and
recognition of objects and their relationships.
Current research in scene graphs focuses on the integration of prior knowledge.
The use of KGs to provide background knowledge and generate scene graphs is recently proposed~\cite{Zareian2020BridgingKG} with Graph Bridging Networks (GB-Net).
The GB-Net regards the scene graph as the image conditioned instantiation of the common sense knowledge graph.
ConceptNet~\cite{Liu2004ConceptNetA} is used as the knowledge graph.
Gu et al.~\cite{Gu2019SceneGG} presented KB-GAN, a knowledge base and auxiliary image generation approach based on external knowledge and image reconstruction loss to overcome the problems found in datasets.
We found no applications of these methods for object detection in AD.

Wickramarachchi et al.~\cite{Wickramarachchi2021} generated a KGE from a scene knowledge graph, and use the embedding to predict missing objects in the scene with high accuracy. 
This is accomplished with a novel mapping and formalization of object detection as a KG link prediction problem. 
Several KGE algorithms are evaluated and compared~\cite{Wickramarachchi2020}. 
Chowdhury et al.~\cite{Chowdhury2021} extended this work by integrating common-sense knowledge into the scene knowledge graph. 
Woo et al.~\cite{Woo2018LinkNetRE} presented a method to embed relations by jointly learning connections between objects. 
It contains a global context encoding and a geometrical layout encoding witch extract global context information and spatial information.

\subsubsection{Road-Sign Detection}
Several approaches focus on using knowledge-graphs for road sign detection~\cite{Kim2020AcceleratingRS,DBLP:conf/semweb/MonkaH0R21}.
Kim et al.~\cite{Kim2020AcceleratingRS} described a method to assist
human annotation of road signs by combining 
KGs and machine learning.
Monka et al.~\cite{DBLP:conf/semweb/MonkaH0R21}
proposed an approach for object recognition based on a knowledge graph neural network (KG-NN) that supervises the training using image-data-invariant auxiliary knowledge.
The auxiliary knowledge is encoded in a KG with respective concepts and their relationships.
These are transformed into a dense vector representation by an embedding method.
The KG-NN learns to adapt its visual embedding space by a contrastive loss function.

\subsubsection{Lane Detection}
Lane detection approaches can be divided into two categories: (1) data from high definition (HD) maps; and (2) data from the vehicle perception sensors (e.g. cameras).
The drawback of HD maps is that such data is not 
always available and not always up to date. 
Traditional methods for lane detection usually perform first edge detection and then model fitting~\cite{Zhu2017OverviewOE}.
A graph-embedding-based approach for lane detection~\cite{Lu2021GraphEmbeddedLD}
can robustly detect parallel, non-parallel (merging or splitting), varying lane width, and partially blocked lane markings.
A novel graph structure is used to represent lane features, lane geometry, and lane topology.
Homayounfar et al.~\cite{Homayounfar2019DAGMapperLT} focused on discovering
lane boundaries of complex highways with many lanes that contain
topology changes due to forks and merges. 
They formulate the problem as inference in a directed \emph{acyclic graph model} (DAG), where the nodes of the graph encode geometric and topological properties of the local regions associated with the lane boundaries.

\subsection{Semantic Segmentation}
The goal of semantic segmentation is to assign a 
semantic meaningful class label (e.g. road, sidewalk, pedestrian, road sign, vehicle) to each pixel in a given image.
A KG-based approach for scene segmentation
is described by Kunze et al.~\cite{Kunze2018ReadingBT}.
A scene graph is generated from a set of segmented 
entities that models the structure of the road using an abstract,
logical representation to enable the incorporation of background knowledge.
Similar approaches based on the (non-semantic) graph representation for scene segmentation can be found in~\cite{Bordes2017EvidentialGA,Dierkes2015TowardsAM,Tpfer2015EfficientRS,Spehr2011HierarchicalSU,Venkateshkumar2015LatentHP}.

\begin{table*}[tb]
  \caption{\textbf{Knowledge Graphs applied to Automated Driving}.}
  \label{tab:KGappliedToAD}
  \small
	\centering
	\begin{tabular}{p{3.9cm}p{4.3cm}p{3.3cm}}
	\hline
  \textbf{Perception} & \textbf{Mapping, Understanding } & \textbf{Plan \& Validate} \\
	\hline 
	\textbf{Object detection} & \textbf{Mapping} & \textbf{Motion Planning} \\
    Scene graphs~\cite{Johnson2015ImageRU,Chang2021SceneGA} &  
        Map representation ~\cite{Suryawanshi2019AnOM}  &  
            Decision making~\cite{Regele2008UsingOT} \\
    Context learning~\cite{Woo2018LinkNetRE}           & 
        Map integration~\cite{Qiu2020AKA} & 
            Rules~\cite{Zhao2015CoreOF,Zhao2015OntologybasedDM,Zhao2017OntologyBasedDD} \\
    KG-scene-graphs~\cite{Zareian2020BridgingKG,Gu2019SceneGG}    & 
        Map updating~\cite{Qiu2020OntologybasedPO} &
            Reasoning~\cite{Huang2019OntologyBasedDS,Xiong2016TheDO} \\
    KG-based detection~\cite{Wickramarachchi2020}    & 
        Quality of maps~\cite{Qiu2020OntologyBasedMD} &
            Rule learning~\cite{Dianov2015GeneratingCM,Hovi2019FeasibilitySR,Morignot2012AnOA} \\
    Common-sense~\cite{Chowdhury2021}  & 
        \textbf{Scene understanding} &
            KG from text~\cite{Elahi2020AFF} \\
    Road sign recog.~\cite{Kim2020AcceleratingRS,DBLP:conf/semweb/MonkaH0R21}& 
        Context model~\cite{Werner2021EmbeddingTS} &
            \textbf{Validation} \\
    Lane detection~\cite{Lu2021GraphEmbeddedLD,Homayounfar2019DAGMapperLT}& 
        Situation understanding~\cite{Halilaj2021AKnowledge} &
            Risk assessm.~\cite{Bagschik2018OntologyBS,Paardekooper2021AHA,Westhofen2022UsingOF} \\
    \textbf{Segmentation} &
       \textbf{Behavior Prediction} & 
            Test gener.~\cite{Chen2018AnOA,Gelder2020OntologyFS,Li2020OntologybasedTG} \\
    Scene segm.~\cite{Kunze2018ReadingBT} & 
       Motion prediction~\cite{Fang2019OntologybasedRA,Zipfl2022Relationa} &
            Verification~\cite{Kaleeswaran2019TowardsIO} \\
	\hline
	\end{tabular}
	\vspace{-0.3cm}
\end{table*}

\subsection{Mapping}
Automated vehicles often use digital maps as a virtual sensor to retrieve 
information about the road network for understanding, navigating, and making decisions about the driving path.
Qui et al.~\cite{Qiu2020AKA} proposed a knowledge architecture with two levels of abstraction to solve the map data integration problem. 
How to use different types of rules to achieve 
two-dimensional reasoning is detailed in~\cite{Qiu2020OntologybasedPO}. 
Qiu et al.~\cite{Qiu2020OntologyBasedMD} addressed the issue of quality assurance in ontology-
based map data for AD, specifically the detection and rectification of map errors.

\subsection{Scene Understanding}
Scene understanding aims to understand what is happening in the scene, the relations between the objects in order
to obtain a comprehension for further steps in automated driving that deal with motion planning and vehicle control.
An approach for KG-based scene understanding was described by Werner et al.~\cite{Werner2021EmbeddingTS}.
The paper proposes a KG to model temporally
contextualized observations and Recurrent Transformers (RETRA), a neural encoder stack with a feedback loop and constrained multi-headed self-attention layers.
RETRA enables transformation of global KG-embeddings into custom embeddings, given the situation-specific factors of the relation and the subjective history of the entity.
Halilaj et al.~\cite{Halilaj2021AKnowledge} presented a KG-based approach for fusing and organizing heterogeneous types and sources of information for driving assistance and automated driving tasks.
The model builds on the terminology of ~\cite{DBLP:conf/itsc/UlbrichMRSM15} and uses
existing ontologies such as SOSA~\cite{Janowicz2019SOSAAL}.
A KGE based on graph neural networks~\cite{Wang2017KnowledgeGE} is then used for
the task to classify driving situations. 

\subsection{Object Behavior Prediction}
Fang et al.~\cite{Fang2019OntologybasedRA} described an ontology-based reasoning approach for long-term
behavior prediction of road users. 
Long-term behavior is predicted with estimated probabilities based on semantic reasoning
that considers interactions among various players. 
Li et al.~\cite{Li2019GRIPGI} present a graph based interaction-aware trajectory prediction (GRIP) approach.
This is based on a GCN and graph operations that model the interaction between the vehicles. 
Relation-based traffic motion prediction using traffic scene graphs and GNN is described in~\cite{Zipfl2022Relationa}.

\subsection{Motion Planing}
The aim of motion planning is to plan and execute driving actions such as steering, accelerating, and braking taking all information of previous steps into account.
Regele~\cite{Regele2008UsingOT} uses an ontology-based 
high-level abstract world model 
to support the decision-making process. 
It consists of a low-level model for trajectory planning and 
a high-level model for solving traffic coordination problems. 
Zhao et al.~\cite{Zhao2015CoreOF,Zhao2015OntologybasedDM,Zhao2017OntologyBasedDD} 
presented core ontologies to enable safe automated driving.
They are combined with rules for ontology-based decision-making on uncontrolled intersections
and narrow roads. 
Another approach for ontology development, focusing on vehicle context is described in~\cite{Xiong2016TheDO}.
Huang et al.~\cite{Huang2019OntologyBasedDS} use ontologies for scene-modeling,
situation assessment and decision-making in urban environments and
a knowledge base of driving knowledge and driving experience. 
Using ontologies to generate semantic rules from a decision tree classifier trained on driving test descriptions of a driving school is outlined in~\cite{Dianov2015GeneratingCM} .
Khan et al.~\cite{Elahi2020AFF} retrieve human knowledge from natural text descriptions of traffic scenes with pedestrian-vehicle encounters. 
Hovi~\cite{Hovi2019FeasibilitySR} presents how rules for 
ontology-based decision-making systems can be learned through machine learning.
Morignot~\cite{Morignot2012AnOA} presents an ontology-based approach for decision-making and relaxing traffic regulations in situations when this is a preferred scenario.

\subsection{Validation}
In the following, we list KG-based approaches that support the validation of AD systems including requirements verification, test case generation, and risk assessment.
Bagschik et al.~\cite{Bagschik2018OntologyBS} introduced a concept of ontology-based scene creation,
that can serve in a scenario-based design paradigm to analyze a system from multiple viewpoints.
In particular, they propose the application for hazard and risk assessment. 
A similar approach for use case generation is described in~\cite{Chen2018AnOA,Gelder2020OntologyFS}.
Paardekooper et al.~\cite{Paardekooper2021AHA} presents a hybrid-AI approach to situational awareness.
A data-driven method is coupled with a KG along with the reasoning capabilities to increase the safety of AD systems.
Criticality recognition using the {A.U.T.O} ontology is demonstrated in~\cite{Westhofen2022UsingOF}.
Li et al.~\cite{Li2020OntologybasedTG} outlined a 
framework for testing, verification, and validation for AD. 
It is based on ontologies for describing the environment and converted to input models for combinatorial testing. 
Kaleeswaran et al.~\cite{Kaleeswaran2019TowardsIO} presented an approach for verification of requirements in AD. 
A semantic model translates requirements using world knowledge into formal representation, which is then used
to check plausibility and consistency of requirements.

A summary of KG-based methods for AD-tasks is shown in Tab.~\ref{tab:KGappliedToAD}. 
It can be seen, that for every component and task only very few KG-based approaches have been developed, suggesting much potential for further research in this area.

\section{Conclusions}
While AD has made tremendous progress over the last few years, many questions remain still unanswered. 
Among these are verifiability, explainability, and safety.
AD systems that operate in complex, dynamic, and interactive environments require approaches that generalize to unpredictable situations and that can reason about the interactions with multiple participants and variable contexts.

Recent progress on KGs and KG-based representation learning has opened new possibilities in addressing these open questions.
This is motivated by two properties of KGs.
First, the ability of KGs to represent complex structured information and in particular, relational information between entities; 
second, the ability to combine heterogeneous sources of knowledge, such as common sense knowledge, rules, or crowd-sourced knowledge, into a unified graph-based representation.
Recent progress in KG-based representation learning opened a new research direction in using KG-based data for machine learning 
applications such as AD. 

We have surveyed ontologies and KG-based approaches for AD.
A few automotive ontologies have been developed. 
However, harmonization of automotive ontologies and 
AD dataset structures is an important step towards enabling KG construction and usage for AD tasks.
Only a few KG-based methods have been developed for AD.
Given the benefits and improvements of KG-based methods
in other domains indicates great potential of 
KG-based methods in the AD domain.

Future topics include but are not limited to (1) the inclusion of additional knowledge sources; (2) task-oriented knowledge representation and knowledge embedding; (3) temporal representation of KG-based approaches; and (4) rule extraction and verification for explainability.
While KGs have already been used in the areas of perception, scene understanding, and motion planning, the use of the technology for the tasks of sensing and act \& control will bring further advances for AD.
The work indicates that knowledge graphs will play an important role in making automated driving better, safer, and ultimately feasible for real-world use.


\bibliographystyle{splncs04}
\bibliography{bibliography}

\end{document}